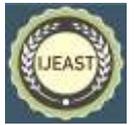

# WEB AND MOBILE PLATFORMS FOR MANAGING ELECTIONS BASED ON IOT AND MACHINE LEARNING ALGORITHMS

Galagoda G.M.I.K, Karunarathne W.M.C.A, Bates R S, Gangathilaka K.M.H.V.P,
Kanishka Yapa, Erandika Gamage
Sri Lanka Institute of Information Technology Malabe, Sri Lanka

*Abstract* — **The global pandemic situation has severely affected all countries. As a result, almost all countries had to adjust to online technologies to continue their processes. In addition, Sri Lanka is yearly spending 10 billion on elections. We have examined a proper way of minimizing the cost of hosting these events online. To solve the existing problems and increase the time potency and cost reduction we have used IoT and ML-based technologies. IoT-based data will identify, register, and be used to secure from fraud, While ML algorithms manipulate the election data and produce winning predictions, weather-based voters' attendance, and election violence. All the data will be saved in cloud computing and a standard database to store and access the data. This study mainly focuses on four aspects of an E-voting system. The most frequent problems across the world in E-voting are the security, accuracy, and reliability of the systems. E- Government systems must be secured against various cyber- attacks and ensure that only authorized users can access valuable, and sometimes sensitive information. Being able to access a system without passwords but using biometric details has been there for a while now, however, our proposed system of ours has a different approach to taking the credentials, processing, and combining the images, reformatting and producing the output, and tracking. In addition, we ensure to enhance e-voting safety. While ML- based algorithms use different data sets and provide predictions in advance.**

## I. INTRODUCTION

The COVID-19 pandemic has taught us a lot of lessons during the last two years. Due to that use of the technology has been improved and increased in various ways. No surprise, when it comes to governing a country and stabilizing democracy it is also digitalized in various ways comparatively faster than before. However, holding elections safely and upholding democracy during the COVID-19 pandemic was one of the biggest challenges that the entire
globe had to face. Especially when it comes to the cyber safety aspect and e-voting. Some European and Latin American countries including the USA and Canada have made a positive approach to governing and e-voting. However, In [4]. according to recent research done by Canada, they have noticed the seriousness of the cyber safety measures that they had to take care of when they executed. Not only the safety aspect of the system but also the e-voter authorization and accessing the details were also seriously taken care of. [3], [7], [8]. Even worse, the financial situation in the country and the considerably large amount of money that the government is spending on executing public voting are extremely high. [9], [10], [11], [12]. A positive result from this project will be able to minimize more than 90% of the cost of the system we suggest.

With this approach, it is a must to develop features to ensure the security aspect of the system. In addition to that when continuing this research, we should be more considerable to focus on the cyber security aspect, user authorization with biometric details, image processing technologies, cloud and database safety, and ethical hacking.

In addition, ensuring to reduce the cost of paper ballot papers and human efforts, this implementation transfers the manual process into a fully automated process. While using E-ballot paper and real-time results calculation. To make things more efficient and to predict the possibilities of an election result we have used a machine learning algorithm with a data set that can produce the election predictions halfway through the election. In advance, we have used weather details and tracked them down with another data set by aligning them with the past election results depending on the weather of different areas to predict the possible vote counts and the public involvement using another ML algorithm. As such, voting violence is a common issue around the world. This system facilitates the admins to monitor the election violence predictability in different areas using another ML- based algorithm.

According to previous research done in this domain, even India has come up with a password method for e-voting [13]. However, that suggested method may cause fraud when voting and arise problems which, apart from that as mentioned above in [4] take huge steps to improve efficient cyber security measures. Moreover, the suggested platform is a multipurpose web application that is platform independent. By doing so, overall cloud safety and the entire system safety depend on this research project.

## II. LITERATURE REVIEW

This section can be mainly divided into several parts according to the different approaches that the authors have been taken

### A. IoT based System authorization and monitoring

[01] We propose a new secure authentication for an online voting system by using biometric features and steganography. A voter is asked to enter a password at the time of registration, which is converted into a secret message using timestamp and hashing. In this model, a person can also vote from outside of his/her allocated electorate or from his/hers chosen location.

[02] In October 2019 federal election promises to be the first in Canadian history where "election cyber security" will play a prominent role. Election cyber security can be understood as preventing digital interference with the main actors, institutions, and processes of elections. This essay outlines the major election cyber security issues facing Canada by focusing on three key actors — namely, political parties, election administrators, and voters.

[03] An election is not a single event but rather a process. It is thus helpful to consider the information technology (IT) of voting in two logically distinct categories: IT for voter registration and IT for voting.

[04] The traditional voting process is quite inconvenient because of the reluctance of voters to visit the booth. The huge evolution in computer technology has invoked us to develop an online voting system that is easy, convenient, and efficient. It is a twofold system consisting of an SMS voting system and a website voting system. A new approach to voting breaks the limitation of traditional voting and focuses on the security and feasibility of voting.

[05] U.S. municipalities and states are adopting paperless electronic voting systems from several different vendors. Our analysis shows that this voting system is far below even the most minimal security standards. Insider threats are not the only concern; outsiders can also do damage. We suggest that the best solutions are voting systems with a voter-verifiable audit trail that can be read and verified by the voter.

[06] An online voting system for the Indian election is proposed for the first time in this paper. The proposed model has greater security in the sense that the voter's high-security password is confirmed before the vote is accepted in the main database of the Election Commission of India. In the proposed system the tallying of the votes will be done automatically, thus saving a huge time and enabling the Election Commissioner of India to announce the result within a very short period.

### B. ML Based vote result prediction

[09] An online voting system for the Indian election is a proposed model that has a high-security password for the voter confirmed before the vote is accepted in the main database of the Election Commission of India. In the proposed system the tallying of the votes will be done automatically, thus saving time and enabling the Election Commissioner of India to announce the result within a very short period. The additional feature of the model is that the voter can confirm if his/her vote has gone to the correct candidate/party.

[10] In India democratic form of government is run by the elected representatives of the common people. Voting is important because the people participate in elections to choose their representatives. Their objective is to design and develop a framework for a secure online voting system in a cloud environment by using a digital certificate secure hash algorithm The proposed system consists of three managers such as certification authority manager (CAM), Vote manager (VM), and vote tally manager (TM). Though there are several issues in implementing an online voting system, security is one of the most important issues. The assessment of the system is verified using properties called authenticity, verifiability, confidentiality, accuracy, and non-reputation.

[11] In this paper, they discussed various voting systems and their advantages and disadvantages. The primary goal of this paper is to make the voting system multipurpose and make it work multiplatform on any operating system.

[12] Manual voting may lead to malpractice sometimes. so there is a need to implement an online voting system. In this specific research, their idea is to implement an online voting system with features like the schemes that the specific party has implemented, based on the features we are going to vote for. They have implemented this by using C# as a programming language, Microsoft SQL server 2012, and Microsoft Azure as a cloud.

[13] Every voting system, either traditional paper voting or electronic voting needs to satisfy the required security properties. This paper renders a survey of various kinds of electronic voting systems with their strengths and defects.

[14] The introduction of technology into voting systems can bring several benefits, such as improving accessibility, remote voting, and efficient, accurate processing of votes. In addition to the undoubted benefits, the introduction of such technology introduces particular security challenges, some of which are unique to voting systems because of their specific nature and requirements.

[15] This paper seeks to discuss the issue of financing of election campaigns within the particular social, political, and legal context of Sri Lanka, and draws parallels and lessons from comparative jurisdictions, with a view to enumerating the options and considerations that must be considered in assessing the need for a regulatory framework.

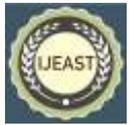

*C.* **Weather-based voter attendance predictor using ML**

[16] To get around the issues, use a revolutionary voting framework-based visual analytics solution. The framework uses the concept of a majority vote to carefully blend the various analog technique variations in order to effectively get the right analogs for calibration. The technology effortlessly incorporates analog techniques into a pipeline of interactive visualization with a collection of coordinated views that represent the various techniques. To assist users in locating and honing analogs, the views offer immediate visual cues. To create the system, we collaborated extensively with subject matter specialists in meteorological research. Two case examples are used to illustrate the system's efficacy. The system's usability and value have been demonstrated by an informal evaluation with the experts.

[17] Data at the state level from the presidential elections in the United States (N = 761). Changes in temperature and voter turnout were positively correlated, which is in line with the excitation transfer theory, which postulates that heat-induced arousal can transfer to other activities and strengthen those activities. Additionally, a rise in temperature was associated with an increase in votes for the ruling party. These results contribute to the body of knowledge about the significance of non-ideological and non-rational factors that affect voting behavior.

[18] Voter behavior can vary with rises in state-level temperature, according to the first study to assess the impact of temperature fluctuations on peaceful and democratic political behavior. Voter turnout climbed by 1.4 percent for every 10C rise in temperature. Additionally, people choose to support the ruling party when the weather is warmer.

[19] Election results and campaign contributions are impacted by extreme weather and natural disasters. Climate change-related weather events may have an impact on these results by prompting people to reconsider the environmental stance of the incumbent leader. In a short-run analysis, we find that the number of online Democratic Party donations rises in response to higher weekly temperatures, with the effect being stronger in counties with more incumbent lawmakers who oppose the environment. In a medium-run analysis, we find that when a natural disaster happens, the election becomes more competitive if the incumbent is more anti-environmental. As a result, the race is more likely to be fought and the incumbent is less likely to win reelection. Election results and campaign contributions are impacted by extreme weather and natural disasters. Climate change-related weather events may have an impact on these results by prompting people to reconsider the environmental stance of the incumbent leader. In a short-run analysis, we find that the number of online Democratic Party donations rises in response to higher weekly temperatures, with the effect being stronger in counties with more incumbent lawmakers who oppose the environment. In a medium-run analysis, we find that when a natural disaster happens, the election becomes more competitive if the incumbent is more anti-environment. As a result, the race is more likely to be fought and the incumbent is less likely to win reelection.

[20] Describe the significance and development of energy policy related to climate change in Swiss elections over the last 15 years. We merged information we had gathered on the issue of energy transition—the main weapon for preventing climate change—in party manifestos with the findings of recent voter preference research conducted in Switzerland. According to our research, voters and political parties are becoming more divided over energy and environmental issues. The populist right SVP was mostly responsible for this development. The importance of the issue has simultaneously increased among voters and political parties on the right of the political spectrum. The space in which parties fight along this dimension is becoming more constrained as issue-owning green parties, however, pay less attention to energy and environmental issues while ignoring the substantial rising trend of these issues among their core supporters.

[21] There has been a significant rise in public awareness of climate change because of the increasing frequency of extreme weather events around the world, the intensifying international discussions about the need for immediate political action to combat climate change and the corresponding more action-oriented social movements. However, there hasn't been any systematic study of how this development affects behavior. We, therefore, examine whether the current rise in climate change awareness has primarily changed public opinions towards environmental and sustainability issues or whether it has resulted in sustainable behavioral shifts using Germany as a case study. We investigated two ways that an increase in climate change awareness could trigger changes based on prior research: (a) directly through influencing behavioral shifts toward more environmentally friendly purchasing decisions or (b) by applying pressure to the political process indirectly.

### III. METHODOLOGY

Mainly the system is a fully integrated web and mobile application-based solution that uses IoT devices when onboarding the users by taking the user credentials and the biometric details. The system will monitor and authenticate the users depending on different data sets when giving access to different election campaigns. In addition, three machine learning algorithms use three different data sets to produce more accurate predictions on winning, attendance, and election violence which may play a major role when hosting and managing such events. The system uses web API to access the DB and transfer the data from users to the DB

This section can be mainly divided into several parts according to the different approaches that the authors have taken.

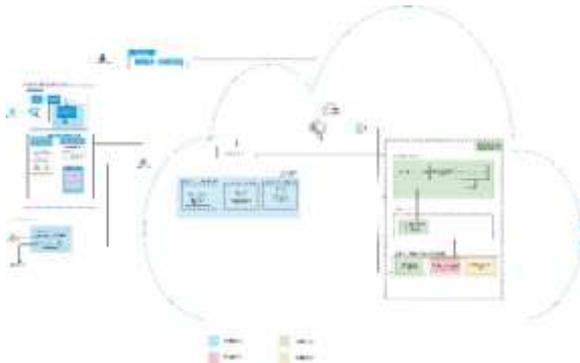

Figure 01: System Diagram

### A. IoT based System authorization and monitoring

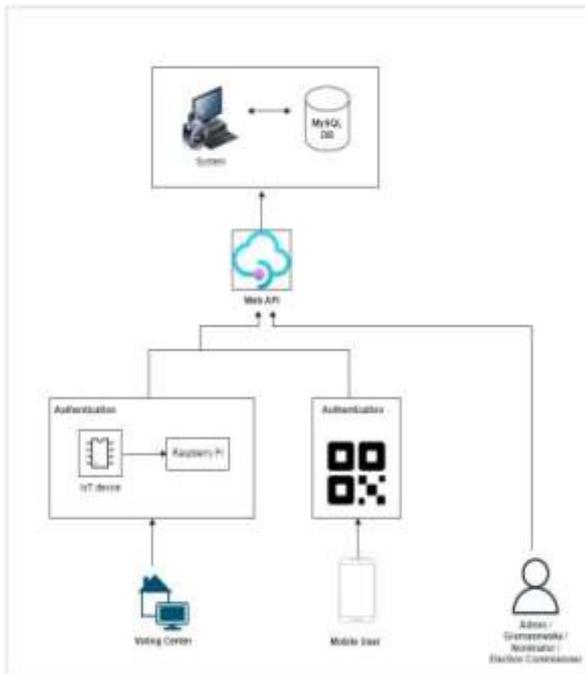

Figure 02: IoT Based System Diagram

According to the above diagram I am responsible for getting the user biometrics, image details, and primary information using a government-authorized person by using an interface defined and allocated by the system. (Web and mobile- based). In addition, an IoT device to take the user's fingerprint has been developed and introduced. The details will be added to the database. Each user will have a separate slot and data will be stored separately. Then the users will have to log in to the system by using their Nic and biometric details. However, at that point, the new biometric details and image details will be compared and if its similar users have access to the system otherwise users will have to start the process again. The gathered data will be stored in a chain. Each person will have separate data storage and access data individually when needed. Moreover, the developed system will be having an election authorization algorithm that will facilitate when the elections are taking place. When the admin users create location-based elections, only the location-based registered voters will be eligible to vote. However, they will be able to either go to any voting station and do the voting or they can use their mobile phones and do the voting. However, we have created a QR method for mobile voters which will be used to authorize the users. Moreover, the system super admin functionalities should be handled through a super admin function. And the super admin functionalities will also be handled and created through web-based technologies.

### B. Weather-based voter attendance predictor using ML

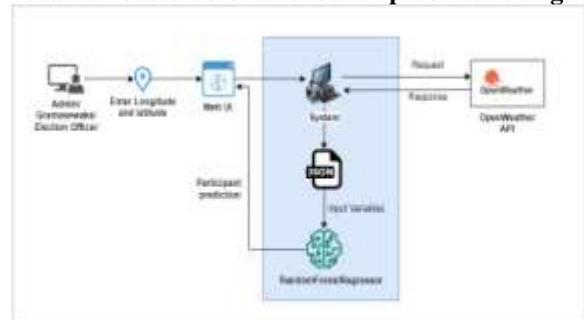

Figure 03: Weather-based voter attendance System Diagram

First Admin / Gramasewaka or the Election Officer logs in to the web application. And redirected to the admin dashboard. Admin enters the longitude and the latitude using the User Interface (UI). After the data inputs, it will pass to the system as two parameters. The request will be sent with the two parameters to the OpenWeather Application. The Open Weather web service offers global weather data through API, including real-time weather information, forecasts, nowcasts, and historical weather information for every place. For any place, the firm offers a minute-by-minute hyperlocal precipitation forecast. So once the API request is sent the response will be sent back from the Open Weather web service to the system. The System will store the response data as variables in a JSON format. The data received as the response are visibility, humidity, temperature, wind Speed, and cloudiness. These variables are passed to the model and the model analyzes the variables and gives a prediction of the participant count.

### C. Election violence predictor using ML

This section describes the process followed to make the e-voting system a reality and the process followed to implement the predictive model for election violations. We have followed the agile methodology throughout the development of the system. Before starting the project, we did research to identify the major problems of the elections in Sri Lanka. We have found that most of the problems are with the high cost and the less security. Then we listed down the major problems in these specific areas and found out that election violation is a major problem in the current election system. As a solution, we decided to develop a model to predict election violations for each area using historical data. From that, the upper management of the election can take decisions according to the high-risk areas.

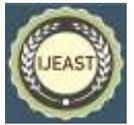

After deciding on the predictive model as the solution, we did a literature survey on the available research. Then we listed the drawbacks and the shortcomings of the selected research. With including the solutions for the found drawbacks we designed the model. Then we did a feasibility study for the election violation prediction. Then we focused on gathering functional and non- functional requirements of the system. After gathering requirements, We developed a high-level architecture diagram of the system. Then I started developing the predictive model for election violation based on the warehouse data for each area.

First, we did a literature survey to identify the best predictive algorithms from predictive models. Then we started the implementations of the automated predictive feature and then started the manual prediction feature. For that we trained the model, then tested and evaluated models. Finally, we implemented some parts of the interactive dashboard as a web application.

## IV. RESULTS AND DISCUSSION

The following material has been separated into four sections for the reader's convenience. As a result, the parts that follow will expound on the mechanism process considering the research. Sections should be read in order.

### A. Data gathering
*a)* IOT
The data gathering part will take place initially when the "Gramasevaka" uses the given application to fill in the information. In this step, we are taking the fingerprint and user images and save them into the DB. User NIC number will work as the primary key while fingerprint and the image details together will be used for authorization

*b)* ML Based Vote result predictor Data sets
We will be collecting the previous voting results sheets and the attendance at the different voting stations during different times of the day.

*c)* Weather-biased voters attendance predictor
We will be collecting the previous voting station attendance data sets and the weather reports of those specific dates in those specific areas.

*d)* Election violence predictor
As a solution, we decided to develop a model to predict election violations for each area using historical data. From that, the upper management of the election can take decisions according to the high-risk areas.

### B. Training and testing data
*a)* IOT
Data will be manipulated using algorithms. This data will be used to specifically identify and access the electoral despite of the voting station or even when using mobile voting.
Testing has been divided into several sprints. where web-based testing, IoT testing, and actual expert testing take place.

*b)* ML Based Vote result predictor Data sets
Testing has been divided into several sprints. where web-based testing, ML testing, and actual expert testing take place.

*c)* Weather-biased voters attendance predictor
When considering the datasets, the main inputs required for this component are visibility, humidity, temperature, wind speed, and cloudiness which contain the requirements that the participant count depends on the above factors. The prepared data set contains 500 records. The figures below show the Data Set prepared to train the model.
Testing has been divided into several sprints. where web-based testing, ML testing, and actual expert testing take place.

*d)* Election violence predictor
Testing has been divided into several sprints. where web-based testing, ML testing, and actual expert testing take place.

### C. Data Pre-processing
*a)* IOT
This will be taking place after all the data has been taken into the DB. From the DB, by using a simple algorithm we will be categorizing the users into area-specific groups which will be used when an election takes place. From the user table, we will be authenticating the users by the biometric process when it comes, they visit the polling stations. For mobile voters, we will be sending an election- specific one-time-only QR code that can be used as authorization to vote.

*b)* ML Based Vote result predictor Data sets
 The predictive model should contain a variety of various elements that have the potential to affect future behavior or outcomes. The first step is to sample the data using a suitable sampling strategy from the data mart or data warehouse. This requires the use of stratified sampling or randomized statistical sampling, such as cluster sampling. The system must produce the Test Set, Training Set, and Holdout Set using those procedures

*c)* Weather-biased voters attendance predictor
Considering the model implementation, we used Random Forest Regression. It is a supervised learning technique called Random Forest Regression that leverages the ensemble learning approach for regression. The ensemble learning method combines predictions from various machine learning algorithms to provide predictions that are more accurate than those from a single model. The tree is built at random throughout the learning process. Every tree contains an internal node where the classification and characteristics of each leaf node are decided.

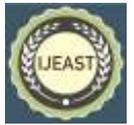

*d)* Election violence predictor

There are three main functions that occur while developing the Predictive Model. They are,
- Data Sampling
- Factor Analysis
- Pattern Recognition

The predictive model should contain a variety of various elements that have the potential to affect future behavior or outcomes. The first step is to sample the data using a suitable sampling strategy from the data mart or data warehouse. This requires the use of stratified sampling or randomized statistical sampling, such as cluster sampling. The system must produce the Test Set, Training Set, and Holdout Set using those procedures

### D. Feature Extraction
*e)* IOT

It is planned to combine IoT and the user image details into one specific image ID that will work as one individual image which will be used in the authentication process while using the user data to be integrated with all the other users.

*f)* ML Based Vote result predictor Data sets

We are planning to use this information on pre result prediction module that can be used to monitor different aspects when it comes to sports and other winning prediction technologies.

*g)* Weather-biased voters attendance predictor

The goal is to advance the system to a point where this can be developed in a way where we can predict people's attendance at different events depending on the weather.

*h)* Election violence predictor

The goal is to improve the system to a point where we can use it to track domestic violence and other sorts of violence where police and other organizations can take precautions in red-light areas.

## V. CONCLUSION AND FUTURE WORK

The proposed system which consists of an automated election management system will facilitate any government in cost reduction, timesaving, and making the process easier while benefiting the ease of access to the community. The system can facilitate the features to minimize the risk of fraud and external hacking and unauthorized actions. While machine learning predicts the important aspects of predictions not only can be used for this system but also for different aspects as well. Such a system has huge potential in the business aspect as well

## VI. ACKNOWLEDGMENT

First and foremost, we want to thank our supervisor for his essential contributions to our research. Thank you to the Sri Lanka Institute of Information Technology for allowing us to work on a research project that allowed us to refresh all of the concepts and technologies we learned during our degree. All parties have been quite helpful, and we appreciate that.

## VII. REFERENCES

[1]. Himanshu Agarwal and G. N. Pandey.( 1 Nov. 2013) "Online Voting System for India Based on AADHAAR ID." IEEE Xplore, [online],Available: https://ieeexplore.ieee.org/document/6756265

[2]. [ Accessed: 24 Jan. 2022]

[3]. S. Vijayalakshmi and G. R. Karpagam.( Jan. 2018) "Secure Online Voting System in Cloud." Electronic Government an International Journal, ResearchGate, [online], Available: www.researchgate.net/publication/326636528_Secure online_voting_system_in_cloud.[ Accessed: 24 Jan. 2022]

[4]. Z.A. Usmani, Kaif Patanwala, Mukesh ,Panigrahi, Ajay Nair.() "Multi-Purpose Platform Independent Online Voting System." IEEE Xplore, [online], Available:https://ieeexplore.ieee.org/document/8276077. [ Accessed: 24 Jan. 2022]

[5]. Ramya Govindaraj, P Kumaresan and K. Sree harshitha.( 1 Feb. 2020) "Online Voting System Using Cloud." IEEE Xplore, [online],Available: https://ieeexplore.ieee.org/document/9077751[ Accessed: 25 Aug. 2021.]

[6]. Htet Ne Oo and Aye Moe Aung.( July 2014,) "A Survey of Different Electronic Voting Systems." ResearchGate,[online],Available: www.researchgate.net/publication/321431416_A_Survey_of_Different_Electronic_Voting_Systems.[ Accessed: 24 Jan. 2022.]

[7]. Y. A. Ryan, P. (2013). Verifiable Voting Systems. [online] ResearchGate. Available at: https://www.researchgate.net/publication/287234012_Verifiable_Voting_Systems

[8]. Sankhitha Gunaratne,( May 2017.) "A Brief on Election Campaign Finance in Sri Lanka", Transparency International Sri Lanka, First Edition [Online]. Available: https://www.tisrilanka.org/wp-content/uploads/2019/05/CampaignFinance.pdf [Accessed: 23 Jan. 2022.]

[9]. Zulfick Farzan,( 23 Jun, 2020.) "2020 General Election Cost to be kept below Rs. 10 Billion:

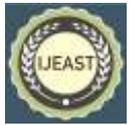

Elections Chief," newsfirst.lk, [Online]. Available:https://www.newsfirst.lk/2020/06/23/2020-general-election-cost-to-be-kept-below-rs-10-billion-elections-chief/ [Accessed: 23 Jan. 2022.]

[10]. H. Liao et al ( 2015) "A visual voting framework for weather forecast calibration," 2015 IEEE Scientific Visualization Conference (SciVis), 2015, pp. 25-32, doi: 10.1109/SciVis.2015.7429488. https://ieeexplore.ieee.org/stamp/stamp.jsp?tp=&arnumber=7429488&isnumber=7429473 [Accessed: October 06th, 2022]

[11]. J Van Assche, A Van Hiel, Stadeus J, Bushman BJ, De Cremer D, Roets (2017 Jun) A. When the Heat Is On: The Effect of Temperature on Voter Behavior in Presidential Elections. Front Psychol. 8;8:929. doi: 10.3389/fpsyg.2017.00929. PMID: 28642723; PMCID: PMC5463178. https://www.ncbi.nlm.nih.gov/pmc/articles/PMC5463178/ [Accessed: October 06th, 2022].

[12]. Frontiers. "Voter behavior influenced by hot weather: Higher voter turnout and an increase in votes for the incumbent political party are linked to state-level increases in temperature during US presidential elections." ScienceDaily. www.sciencedaily.com/releases/2017/08/170816100216.htm [Accessed October 14, 2022].

[13]. Yanjun Liao, Pablo Ruiz Junco, Extreme weather and the politics of climate change: "A study of campaign finance and elections" , Journal of Environmental Economics and Management,Volume 111, 2022, 102550, ISSN 0095-0696, https://doi.org/10.1016/j.jeem.2021.102550. https://www.sciencedirect.com/science/article/pii/S0095069621001078 [Accessed October 06, 2022].

[14]. Lüth, M., & Schaffer, L. M. (2022). "The electoral importance and evolution of climate related energy policy: evidence from Switzerland". Swiss Political Science Review, 28, 169– 189. https://doi.org/10.1111/spsr.12520 [Accessed October 14, 2022].

[15]. Venghaus, S., Henseleit, M. & Belka, M. "The impact of climate change awareness on behavioral changes in Germany: changing minds or changing behavior?". Energ Sustain Soc **12**, 8 (2022). https://doi.org/10.1186/s13705-022-00334- [Accessed October 14, 2022].

[16]. Wipulasena, A., 2022. Sri Lanka election: Observers report poll day violations. [online] Aljazeera.com. Available at: <https://www.aljazeera.com/news/2019/11/16/sri-lanka-election-observers-report-poll-day-violations> [Accessed 16 November 2019].

[17]. HAMZA, M., 2022. CMEV reports 70 violations of election laws as Sri Lanka goes to the polls | EconomyNext. [online] EconomyNext. Available at: <https://economynext.com/cmev-reports-70-violations-of-election-laws-as-sri-lanka-goes-to-the-polls-72660/> [Accessed 5 August 2020].

[18]. Goel, D., Bhatia, R. and Bhatia, K., 2021. Traffic Violations Prediction System on the Basis of Human Behaviour. [online] researchgate. Available at: <https://www.researchgate.net/publication/352635639_Traffic_Violations_Prediction_System_on_the_Basis_of_Human_Behaviour> [Accessed 19 June 2022].

[19]. Sujadmiko, B., P Panggar, I., Sofyansah, A. and Fitri Meutia, I., 2020. The Concept of E-Voting Mechanism Based on Law of General Election and Information Security. [online] Researchgate. Available at: <https://www.researchgate.net/publication/352634389_The_Concept_of_E-Voting_Mechanism_Based_on_Law_of_General_Election_and_Information_Security> [Accessed November 2020].

[20]. Veit, D. and Huntgeburth, J., 2014. E-Voting. [ online] E-Voting. Available at: <https://www.researchgate.net/publication/299688721_E-Voting> [Accessed July 2014].

[21]. Ajibola, O., 2021. E-voting vs Traditional Paper-based voting system. [online] E-voting vs Traditional Paper-based voting system. Available at: <https://www.researchgate.net/publication/364087926_E-voting_vs_Traditional_Paper-based_voting_system> [Accessed December 2021].